\documentclass[10pt,twocolumn,letterpaper]{article}
\pdfoutput=1
\usepackage{cvpr}
\usepackage{times}
\usepackage{epsfig}
\usepackage{graphicx}
\usepackage{amsmath}
\usepackage{amssymb}
\usepackage{authblk}
\usepackage{xcolor,soul}

\usepackage{caption}

\cvprfinalcopy 

\def\cvprPaperID{462} 

\ifcvprfinal\fi
\begin{document}


\title{Watch What You Just Said: \\
Image Captioning with Text-Conditional Attention}

\author[1]{Luowei Zhou}
\author[2]{Chenliang Xu}
\author[3]{Parker Koch}
\author[3]{Jason J. Corso}
\affil[1]{Robotics Institute, University of Michigan}
\affil[2]{Department of Computer Science, University of Rochester}
\affil[3]{Electrical and  Computer Engineering, University of Michigan \authorcr
\tt\small{\{luozhou, pakoch\}@umich.edu}, chenliang.xu@rochester.edu, jjcorso@eecs.umich.edu}

\maketitle

\begin{abstract}
Attention mechanisms have attracted considerable interest in image captioning due to its powerful performance. However, existing methods use only visual content as attention and whether textual context can improve attention in image captioning remains unsolved. To explore this problem, we propose a novel attention mechanism, called \textit{text-conditional attention}, which allows the caption generator to focus on certain image features given previously generated text. To obtain text-related image features for our attention model, we adopt the guiding Long Short-Term Memory (gLSTM) captioning architecture with CNN fine-tuning. Our proposed method allows joint learning of the image embedding, text embedding, text-conditional attention and language model with one network architecture in an end-to-end manner. We perform extensive experiments on the MS-COCO dataset. The experimental results show that our method outperforms state-of-the-art captioning methods on various quantitative metrics as well as in human evaluation, which supports the use of our text-conditional attention in image captioning.
\end{abstract}
 






\section{Introduction}

Image captioning is drawing increasing interest in computer vision and machine learning~\cite{KuPrOrTPAMI2013,MaXuYaICLR2015,YoJiWaCVPR2016,WuShLiCVPR2016}. Basically, it requires machines to automatically describe the content of an image using an English sentence. While this task seems obvious for human-beings, it is complicated for machines since it requires the language model to capture various semantic information within an image, such as objects' motions and actions. Another challenge for image captioning, especially for generative models, is that the generated output should be human-like natural sentences.

Recent literature in image captioning is dominated by neural network-based methods~\cite{DoAnGuCVPR2015, ViToBeCVPR2015, YoJiWaCVPR2016, JiGaFeICCV2015}. The idea originates from the encoder-decoder architecture in Neural Machine Translation~\cite{BaChBeICLR2015}, where a convolutional neural network (CNN) is adopted to encode the input image into a feature vector, and a sequence modeling approach (e.g., Long Short-Term Memory (LSTM)~\cite{HoScNECO1997}) decodes the feature vector into a sequence of words~\cite{ViToBeCVPR2015}. Most recent work in image captioning relies on this structure, and leverages image guidance~\cite{JiGaFeICCV2015}, attributes~\cite{YoJiWaCVPR2016} or region attention~\cite{XuBaKiICML2015} as the extra input to LSTM decoder for better performance. The intuition comes from visual attention, which has been known in Psychology and Neuroscience for a long time~\cite{DeDuARN1995}. For image captioning, this means the image guidance to the language model should change over time according to the context.

\begin{figure}[t]
\centering
\includegraphics[width=\linewidth]{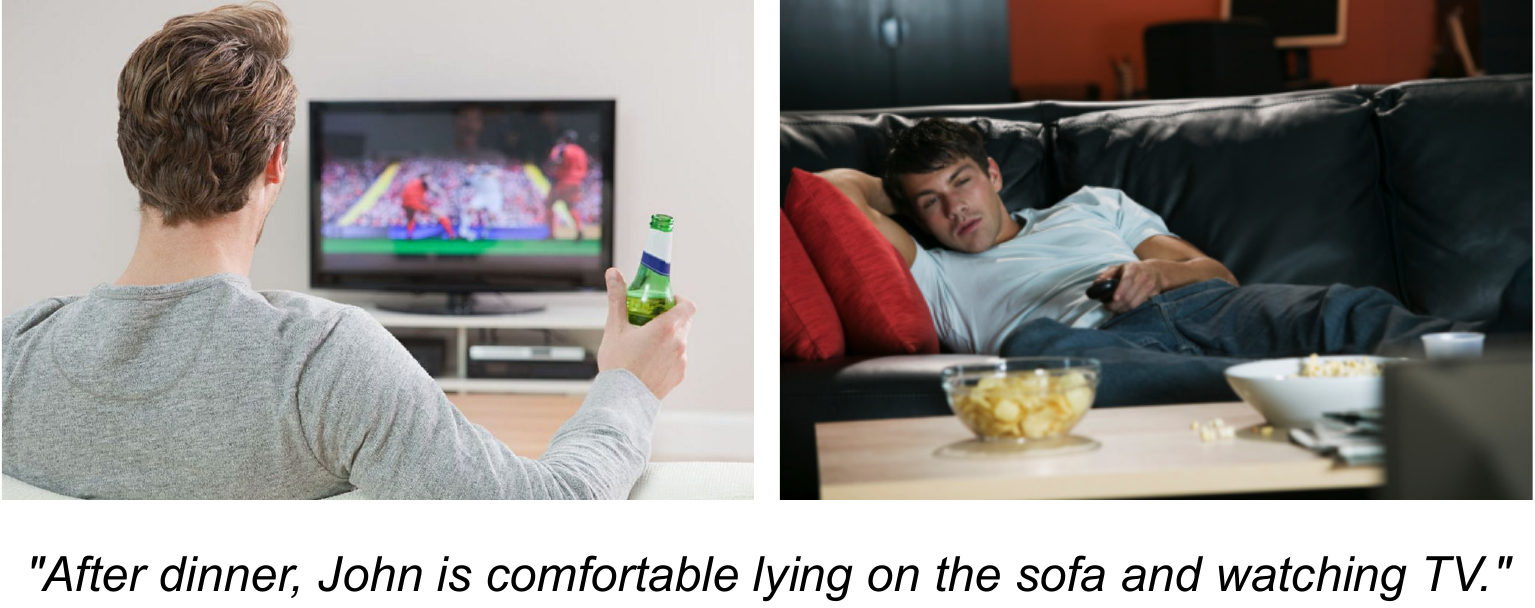}
\caption{We can barely see the \textit{sofa} from the image on the left, and we can only see a corner of the \textit{TV} in the image on the right, but we can infer them from the textual context even with the weak visual evidences}
\label{fig:sofatv}
\end{figure}


However, these methods using attention lack consideration from the following two aspects. First, attending to the image is only half of the story; watching what you just said comprises the other half. In other words, \textbf{visual evidence can be inferred and interpreted by textual context}, especially when the visual evidence is ambiguous. For example, in the sentence ``After dinner, John is comfortably lying on the sofa and watching TV'', the objects ``sofa'' and ``TV'' are naturally inferred even with weak visual evidences (see Figure~\ref{fig:sofatv}, image credits: \url{http://www.mirror.co.uk/} and \url{http://newyork.cbslocal.com/}). Despite its importance, textual context was not a topic of focus in attention models. Existing attention based methods such as~\cite{YoJiWaCVPR2016,WuShLiCVPR2016,XuBaKiICML2015} have used implicit text-guiding from an LSTM hidden layer to determine which of the image regions or attributes to attend on. However, as we mentioned in the previous example, the object for attention might be only partially observable, so the attention input could be misleading. This is not the case for our attention model since the textual features are tightly coupled with the image features to compensate for one another. While Jia et al.~\cite{JiGaFeICCV2015} use joint embedding of the text and image as the guidance for the LSTM, their approach has pre-specified guidance that is fixed over time and has a linear form.  In contrast, our method systematically incorporates the time-dependent text-conditional attention, from 1-gram to n-gram and even to the sentence level.


Second, existing attention based methods separate CNN feature learning (trained for a different task, i.e. image classification) from the LSTM text generation. This leads to a representational disconnect between features learned and text generated. For instance, the attention model proposed by You et al.~\cite{YoJiWaCVPR2016} uses weighted-sum visual attributes to guide the image captioning, while the attributes proposed by the specific predictor are separated from the language model. This makes the attributes guidance lack the ability to adapt to the textual context, which ultimately compromises the end-to-end learning ability the paper claimed. 

To overcome the above limitations, we propose a new text-conditional attention model based on the time-dependent gLSTM. Our model has the ability to interpret image features based on textual context and it is end-to-end trainable. The model learns a text-conditional embedding matrix between CNN image features and previously generated text. Given a target image, the proposed model generates LSTM guidance by directly conditioning the image features on the current textual context. The model hence learns how to interpret image features given the textual content it has recently generated. If it conditions the image features on one previous word, it is a 1-gram word-conditional model. If it is on previous two words, we get a 2-gram word-conditional model. Similarly we can construct an n-gram word-conditional model. The extreme version of our text-conditional model is the sentence-conditional model, which takes advantage of all the previously generated words.

We implement our model \footnote{\url{https://github.com/LuoweiZhou/e2e-gLSTM-sc}} based on NeuralTalk2, an open-source implementation of Google NIC \cite{ViToBeCVPR2015}. We compare our methods with state-of-the-art methods on the commonly used MS-COCO dataset \cite{LiMaBeECCV2014} with publicly available splits \cite{NT2} of training, validation and testing sets. We evaluate methods on standard metrics as well as human evaluation. Our proposed methods outperform the state-of-the-art approaches across different evaluation metrics and yield reasonable attention outputs.

The main contributions of our paper are as follows. First, we propose text-conditional attention which allows the language model to learn text-specified semantic guidance automatically. The proposed attention model learns how to focus on parts of the image feature given the textual content it has generated. Second, the proposed method demonstrates a less complicated way to achieve end-to-end training of attention-based captioning model, whereas state-of-the-art methods~\cite{JiGaFeICCV2015, XuBaKiICML2015, YoJiWaCVPR2016} involve LSTM hidden states or image attributes for attention, which compromises the possibility of end-to-end optimization.

\section{Related Work}

Recent successes of deep neural networks in machine translation~\cite{SuViLeNIPS2014, ChVaGuEMNLP2014} catalyze the adoption of neural networks in solving image captioning problems. Early works of neural network-based image captioning include the multimodal RNN~\cite{KaFeCVPR2015} and LSTM~\cite{ViToBeCVPR2015}. In these methods, neural networks are used for both image-text embedding and sentence generating. Various methods have shown to improve performance with region-level information~\cite{FaGuIaCVPR2015, JoKaFeCVPR2016}, external knowledge \cite{AnVeRoCVPR2016}, and even from question-answering \cite{LiPaECCV2016}. Our method differs from them by considering attention from textual context in caption generating. 

Attention mechanism has recently attracted considerable interest in LSTM-based image captioning~\cite{XuBaKiICML2015, JiGaFeICCV2015, YoJiWaCVPR2016, WuShLiCVPR2016}. Xu et al.~\cite{XuBaKiICML2015} propose a model that integrates visual attention through the hidden state of LSTM model. You et al.~\cite{YoJiWaCVPR2016} and Wu et al.~\cite{WuShLiCVPR2016} tackle the semantic attention problem by fusing visual attributes extracted from images with the input or the output of LSTM. Even though these approaches achieve state-of-the-art performance, the performances rely heavily upon the quality of the pre-specified visual attributes, i.e., better attributes usually lead to better results. Our method also uses attention mechanism, but we consider the explicit time-dependent text attention and is comprised a clean architecture for the ease of end-to-end learning. 

Early works in image captioning focus on either template-based methods or transfer-based methods. Template-based methods \cite{KuPrOrTPAMI2013, LiKuBeCNLL2011, YaTeDaEMNLP2011, MiHaDoEACL2012, ElKeEMNLP2013, GuKrMaICCV2013} specify templates and fill them with detected visual evidences from target images. In Kulkarni et al. \cite{KuPrOrTPAMI2013}, visual detections are first put into a graphical model with higher order potentials from text corpora to reduce noise, then converted to language descriptions based on pre-specified templates. In Yang et al. \cite{YaTeDaEMNLP2011}, a quadruplet consisting of noun, verb, scene and
preposition is used to describe an image. The drawback of these methods is that the descriptions are not vivid and human-crafted templates do not work for all images. Transfer-based methods \cite{FaHeSaECCV2010, KuOrBeACL2012, DaSrCoWSDM2013} rely on image retrieval to assign the target image with descriptions of similar images in the training set. A common issue is that they are less robustness to unseen images.

\subsection{Background}
\label{related:bk}

\begin{figure}
  \centering
  \includegraphics[width=\linewidth]{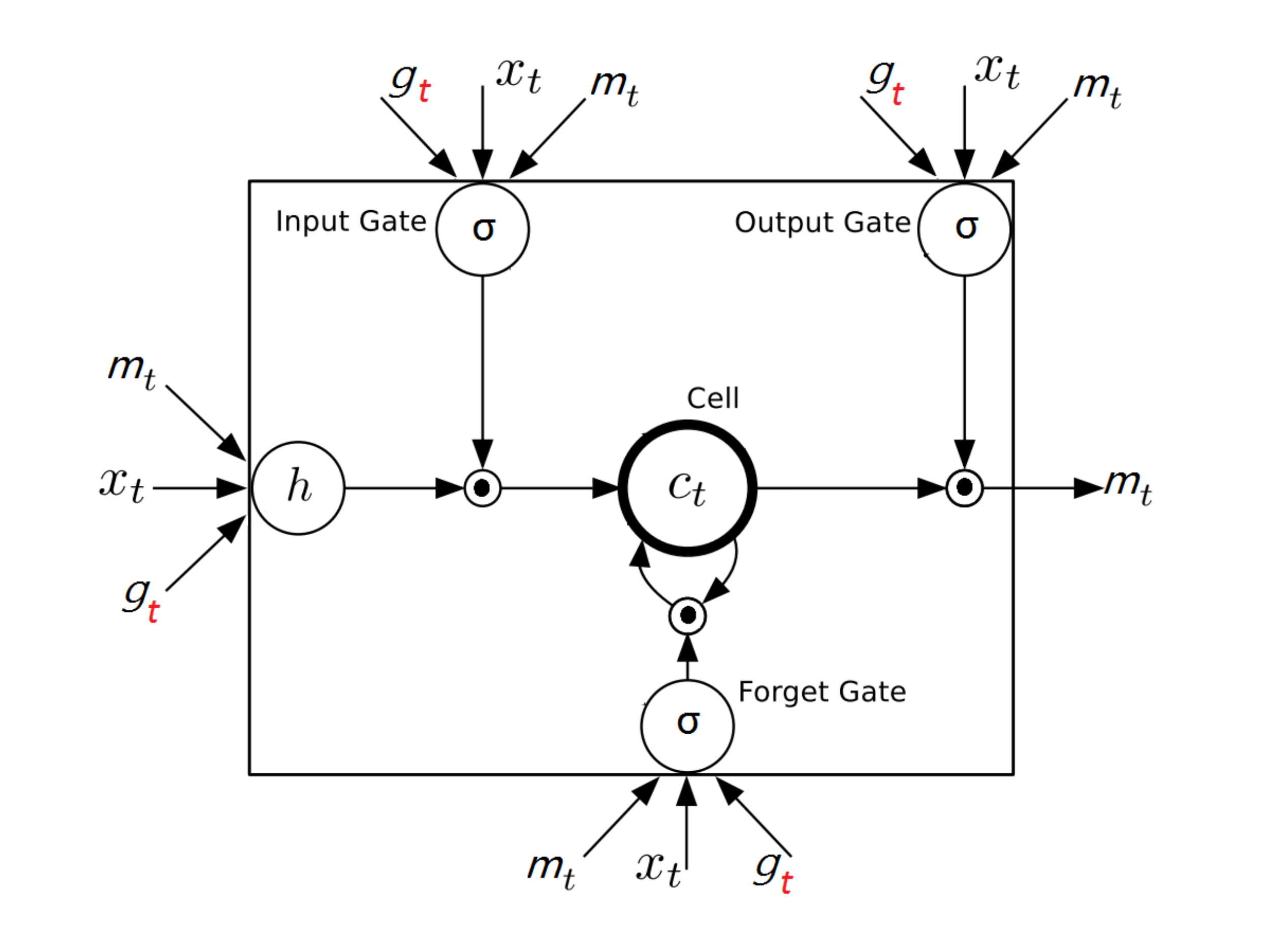}
  \captionof{figure}{Shown in black is the gLSTM node~\protect\cite{JiGaFeICCV2015}. The \textcolor{red}{red} subscript of $g_t$ represents our td-gLSTM guidance. Figure modified from \protect\cite{graves2008super}. Best viewed in color.}
  \label{gLSTM}
\end{figure}

The generated sentences by the LSTM model may lose track of the original image content since it only accesses the image content once at the beginning of the learning process, and forgets the image after even a short period of time. Therefore, Jia et al.~\cite{JiGaFeICCV2015} propose an extension of the LSTM model, named the guiding LSTM (gLSTM), which extracts semantic information from the target image and feeds it into the LSTM model every time step as extra information. The basic gLSTM unit is shown in Fig.~\ref{gLSTM}. Its memory cell and gates are defined as follows:
\begin{align}
&i_t = \sigma(W_{ix}x_t+W_{im}m_{t-1}+W_{iq}g) \nonumber \\
&f_t = \sigma(W_{fx}x_t+W_{fm}m_{t-1}+W_{fq}g) \nonumber \\
&o_t = \sigma(W_{ox}x_t+W_{om}m_{t-1}+W_{oq}g) \nonumber \\
&c_t = f_t\odot c_{t-1}+i_t\odot h(W_{cx}x_t+W_{cm}m_{t-1}+W_{cq}g) \nonumber \\
&m_t = o_t\odot c_t
\enspace,
\end{align}
where $W$s denote weights, $\odot$ represents element-wise multiplication, $\sigma(\cdot)$ is the sigmoid function, $h(\cdot)$ is the hyperbolic tangent function, $x_t$ stands for input, $i_t$ for the input gate, $f_t$ for the forget gate, $o_t$ for the output gate, $c_t$ for state of the memory cell, $m_t$ for the hidden state (also output for one-layer LSTM), and $g$ represents guidance information, which is time-invariant. The subscripts denote time: $t$ is the current time step and $t-1$ is the previous time step.

\begin{figure*}[t]
  \centering
  \centering
  \includegraphics[width=0.8\linewidth]{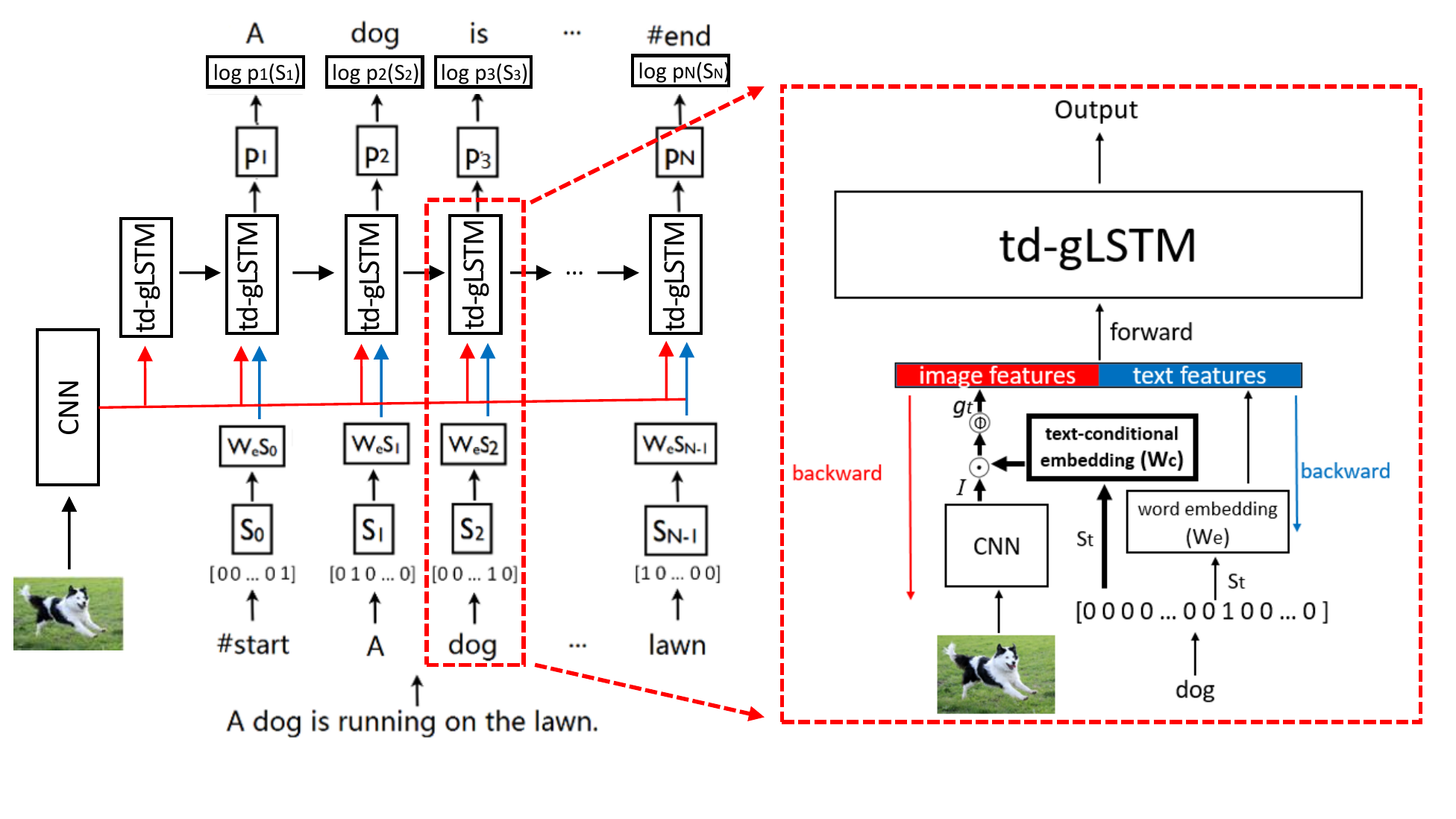}
  \captionof{figure}{Text-conditional semantic attention. The text-conditional attention part are highlighted in \textbf{bold}. $S_t$ indicates the one-hot vector representation of the $t^{th}$ word in the sentence. $W_e$ is word embedding matrix, $W_e$ is text-conditional embedding matrix, \textit{I} is image feature and $g_t$ is time-dependent guidance. See text for more details.}
  \label{fig:tc}
\end{figure*}

\section{Methods}

Our text-conditional attention model is based on a time-dependent gLSTM (td-gLSTM). We first describe the td-gLSTM in Sec.~\ref{sec:td-gLSTM} and show how to obtain semantic guidance through this structure. Then, we introduce our text-conditional attention model and its variants, e.g. $n$-gram word- and sentence-conditional models, in Sec.~\ref{sec:tc}.

\subsection{Time-Dependent gLSTM (td-gLSTM)}
\label{sec:td-gLSTM}


The gLSTM described in Sec.~\ref{related:bk} has a time-invariant guidance. In Jia et al.~\cite{JiGaFeICCV2015}, they show three ways of using such guidance, including an embedding of the joint image-text feature by linear CCA. However, the textual context in a sentence is constantly changing while the caption generator is generating the sentence. Obviously, we need the guidance to evolve over time, and hence we propose td-gLSTM. Notice that, despite its simple change in structure, the td-gLSTM is much more flexible in the way it incorporates guidance, e.g. a time-series dynamic guidance such as tracking and actions in a video. Also, notice that the gLSTM is a sepcial case of the td-gLSTM, when the guidance is set as $g_t = g_{t-1}$. 

Our proposed td-gLSTM consists of three parts: 1) image embedding; 2) text embedding; and 3) LSTM language model. Figure~\ref{fig:tc} shows an overview for using td-gLSTM for captioning. First, image feature vector $I$ is extracted using CNN and each word in the caption is represented by a one-hot vector $S_t$, where $t$ indicates the index of the word in the sentence. We use the text embedding matrix $W_e$ to embed text feature $S_t$ into a latent space, which is the input $x_t$ of the LSTM language model. The text embedding matrix is initialized from a zero-mean Gaussian distribution with standard deviation 0.01. On the other hand, the text feature is jointly embedded with the image feature, denoted as $g_t=h(S_t, I)$, where $g_t$ is the time-dependent guidance. Here, we do not specify the particular form of $g_t$ to make the framework general, and its choices are discussed in Sec.~\ref{sec:tc}.

Both the guidance $g_t$ and embedded text features $x_t=W_e S_t$ are used as the inputs to td-gLSTM, which are shown in Fig.~\ref{gLSTM} (including red) and formulated as follows:
\begin{align}
&i_t = \sigma(W_{ix}x_t+W_{im}m_{t-1}+W_{iq}g_t) \nonumber \\
&f_t = \sigma(W_{fx}x_t+W_{fm}m_{t-1}+W_{fq}g_t) \nonumber \\
&o_t = \sigma(W_{ox}x_t+W_{om}m_{t-1}+W_{oq}g_t) \nonumber \\
&c_t = f_t\odot c_{t-1}+i_t\odot h(W_{cx}x_t+W_{cm}m_{t-1}+W_{cq}g_t) \nonumber \\
&m_t = o_t\odot c_t
\enspace.
\end{align}

We back-propagate error through guidance $g_t$ for fine-tuning the CNN. One significant benefit of this is that the model allows the guidance information to be more similar to its corresponding text description. Note that the text-conditional guidance $g_t$ keeps changing in each time step, which is a time-dependent variable. The outputs of the language model are the log likelihood of each word from the target sentence, followed by a Softmax function for normalization. We use the regularized cross-entropy loss function:
\begin{align}
L(\mathcal{I},\mathcal{S}) = -\sum_{k=1}^{N}\log(p_k(S_k))+\frac{\lambda}{2}||W_{conv}||_2^2
\enspace,
\label{eq:lf}
\end{align}
where $\mathcal{I}$ represents the image, $\mathcal{S}$ represents the sentence, $S_k$ denotes the $k^{th}$ word in the sentence, $S_N$ is the stop sign, $W_{conv}$ denotes all the weights in the convolutional net and $\lambda$ controls the importance of the regularization term. Finally, we back-propagate the loss to LSTM language model, the text embedding matrix and the image embedding CNN. The training detail is described in Sec.~\ref{sec:exp:setup}.


\subsection{Text-Conditional Attention}
\label{sec:tc}

Recently, You et al.~\cite{YoJiWaCVPR2016} use visual attributes as the semantic attention to guide the image captioning. Their semantic guidance consists of top visual attributes of the input image, and the weight of each attribute is determined by the current word, which is the previous output of RNN. However, the attribute predictor adopted in their model has no learning ability and is separated from the encoder-decoder language model. In contrast, following the td-gLSTM model (see Sec.~\ref{sec:td-gLSTM}), we condition the guidance information $g_t$ on the current word $S_t$ (the one-hot vector representation), and use the text-conditional image feature as the semantic guidance. The benefits are twofold: first, the model can learn which part of the semantic image feature should be focused on when seeing a specific word; second, this structure is end-to-end tunable such that the CNNs weights are tuned for captioning rather than for image classification~\cite{russakovsky2015imagenet}. For instance, when the caption generator generated a sequence as ``a woman is washing'', its attention on the image feature should be automatically switched to objects that can be washed, such as clothes and dishes. 

We first consider modeling the text-conditional guidance feature $g_t$ as the weighted-sum of the outer product of image feature $I$ and text feature $S_t$, therefore each entry in $g_t$ is represented as:
\begin{align}
g_{t}^i = \sum_{j,k}W_{ijk}I^{j}S_{t}^k+b^i
\enspace,
\label{eq:3D_tensor}
\end{align}
where $I^j$ denotes the $j^{th}$ entry of the image feature, $S_{t}^k$ denotes the $k^{th}$ entry of the text feature, and $g_{t}^i$ is the $i^{th}$ entry of the text-conditional guidance feature. For each $g_{t}^i$, the corresponding weights $W_i$ is a 2-D tensor, hence, the total weights $W$ for $g_t$ is a 3-D tensor. In this model, image feature is fully coupled with text feature though the 3-D tensor.

Despite Eq.~\ref{eq:3D_tensor} fully couples the two types of features, it results in a huge amount of parameters, which prohibits its use in practice. To overcome it, we introduce an embedding matrix $W_c$, which contains various text-to-image masks. Furthermore, in practice, adding one non-linear transfer function layer after the image-text feature embedding boosts the performance. Therefore, we model the text-conditional feature $g_t$ as a text-based mask on image feature $I$ followed by a non-linear function:
\begin{align}
g_t = \Phi(I\odot W_c S_t)
\enspace,
\label{eq:tc}
\end{align}
where $W_c$ is the text-conditional embedding matrix and $\Phi(\cdot)$ is a non-linear transfer function. When $W_c$ is a all-one matrix, the conditioned feature $I \odot W_c S_t$ is identical to $I$. We transfer the pre-trained model from gLSTM to initialize the CNN, language model and word embedding of our attention model. For text-conditional matrix, we initialize it with all ones. We show the sensitivity of our model to various transfer functions in Sec.~\ref{sec:exp:var}.


The above model is the 1-gram word-conditional semantic attention owing to the guidance feature is merely conditioned on the previous word. Similarly, we develop the 2-gram word-conditional model, which utilizes previous two words, or even n-gram word-conditional model. The extreme version of the text-conditional model is the sentence-condition model, which takes advantage of all the previously generated words:
\begin{align}
g_t = \Phi(I\odot W_c \sum_{k=1}^t \frac{S_{k-1}}{t})
\enspace.
\label{eq:sc}
\end{align}
One benefit of the text-conditional model is that it allows the language model to learn semantic attention automatically though the back-propagation of the training loss while attribute-based method, such as~\cite{YoJiWaCVPR2016}, represents semantic guidance by some major components of an image, but other semantic information, such as objects' motions and locations, are discarded.

\section{Experiments}
\label{sec:exp}

We describe our experiment settings in Sec.~\ref{sec:exp:setup}, analyze the variants of our model and attention in Sec.~\ref{sec:exp:var}, and compare our method with state-of-the-art methods in Sec.~\ref{sec:exp:compare}. 

\subsection{Experiment Setup}
\label{sec:exp:setup}

We use the MS-COCO dataset~\cite{LiMaBeECCV2014} with the commonly adopted splits as described in~\cite{NT2}: 113,287 images for training, 5,000 images for validation and 5,000 images for testing. Three standard evaluation metrics, e.g. BLEU, METEOR and CIDER, are used in addition to human evaluation. We implement our model based on the NeuralTalk2~\cite{NT2}, which is an open source implementation of~\cite{ViToBeCVPR2015}. We use three different CNNs in our experiments, e.g. 34-layer and 200-layer ResNets~\cite{HeZhReCVPR2016} and 16-layer VGGNet~\cite{SiZiICLR2015}. For a fair comparison, we use 34-layer ResNet when analyzing the variants of our models in Table~\ref{tbl:ngram} and~\ref{tbl:tf}, 16-layer VGGNet when comparing to state-of-the-art methods in Table~\ref{tbl:result} and~\ref{tbl:he}, and 200-layer ResNet for leadboard competition in Table~\ref{tbl:official}. The variation of performance regarding different CNNs are also evaluated in Table~\ref{tbl:resnet}.  

We train our model in three steps: 1) train time-invariant gLSTM (ti-gLSTM) without CNN fine-tuning for 100,000 iterations; 2) train ti-gLSTM with CNN fine-tuning for 150,000 iterations; and 3) train td-gLSTM with initializd text-conditional matrix but without CNN fine-tuning for 150,000 iterations. The reason for this multiple-step training is described in Vinyals et al.~\cite{vinyals2016show}: jointly training the system at the initial time causes noise in the initial gradients coming from LSTM that corrupts the CNN unrecoverably. For the hyper-parameters, we set the CNN weight decay rate ($\lambda$ in Eq. \ref{eq:lf}) to $10^{-3}$ to avoid overfitting. The learning rate for CNN fine-tuning is set to $10^{-5}$ and the learing rate for language model is set to $4\times10^{-4}$. We use Adam optimizer~\cite{kingma2014adam} for updating weights with $\alpha=0.8$ and $\beta=0.999$. We adopt $2$ and $3$ for beam sizes during inference, as recommended by recent studies~\cite{vinyals2016show, DoAnGuCVPR2015}. The whole training process takes about one day on a single NVIDIA TITAN X GPU.

\begin{table}[t]
\centering
\caption{Results of n-gram word- and sentence-conditional models with Tanh transfer function and 34-layer ResNet. Top-2 scores for each metric are highlighted. All values are reported as percentage (\%).}
\label{tbl:ngram}
\begin{tabular}{r|ccc}
\hline
 & \textbf{BLEU@4} & \textbf{METEOR} & \textbf{CIDEr} \\
 \hline
\textbf{1-gram} & 29.5 & 24.6 & 94.6 \\
\textbf{2-gram} & 30.2 & 24.8 & \textbf{97.3} \\
\textbf{3-gram} & 29.9 & \textbf{24.9} & 96.1 \\
\textbf{4-gram} & \textbf{30.3} & \textbf{24.9} & 97.0 \\
\textbf{sentence} & \textbf{30.6} & \textbf{25.0} & \textbf{98.1}\\
\hline
\end{tabular}
\end{table}

\begin{table}[t]
\centering
\caption{Results of different transfer functions on sentence-conditional model (denoted as \textbf{sc}) with 34-layer ResNet. Top scores for each metric are highlighted. All values are reported as percentage (\%).}
\label{tbl:tf}
\begin{tabular}{r|ccc}
\hline
& \textbf{BLEU@4} & \textbf{METEOR} & \textbf{CIDEr} \\
\hline
\textbf{sc-relu} & 30.5 & \textbf{25.0} & \textbf{98.1}\\
\textbf{sc-tanh} & \textbf{30.6} & \textbf{25.0} & \textbf{98.1}\\
\textbf{sc-softmax} & 30.2 & 24.9 & 97.1\\
\textbf{sc-sigmoid} & 30.1 & 24.8 & 96.2\\
\hline
\end{tabular}
\end{table}

\subsection{Model Variants \& Attention}
\label{sec:exp:var}

\noindent\textbf{N-gram v.s. Sentence.} Table~\ref{tbl:ngram} shows results with n-gram word- and sentence-conditional models. For conciseness, we only use BLEU@4, METEOR and CIDEr as the evaluation metrics, since they are more correlated with human judgments compared with low-level BLEU scores \cite{vedantam2015cider}. It turns out generally, word-conditional models with higher grams yield better results, especially for METEOR. Notice that the $2,3,4$-gram models achieve considerablely better results than 1-gram model, which is reasonable as the 1-gram has the least context that limits the attention performance. Furthermore, the sentence-conditional model outperforms all word-conditional models in all metrics, which shows the importance of long-term word dependency in attention modeling.

\noindent\textbf{Transfer Function.}  We use a non-linear transfer function $\Phi(\cdot)$ in our attention model (see Eq.~\ref{eq:tc}) and we test four different functions: Softmax, ReLU, Tanh and Sigmoid. In all cases, we initialize the text-conditional embedding matrix with noises from one-mean Gaussian distribution with standard deviation $0.001$. We base our experiments on the sentence-conditional model and conclude that the model achieves best performance when $\Phi(\cdot)$ is a Tanh or a ReLU function (see Table~\ref{tbl:tf}). Notice that it is possible that other transfer functions different than the four we tested may lead to better results. 

\noindent\textbf{Image Encoding.} We study the impact of image encoding CNNs on captioning performance, as shown in Table~\ref{tbl:resnet}. In general, the more sophisticated image encoding architecture the higher performance of the captioning. 

\begin{table}[t]
\begin{center}
\caption{The Impact of image encoding CNNs on captioning performance. Tanh is used as transfer function. Top result for each column is highlighted. All values are reported as percentage (\%).} 
\renewcommand\arraystretch{1.2}
\label{tbl:resnet}  
\begin{tabular}{r|ccc} 
\hline
\textbf{Methods} & \textbf{BLEU@4} & \textbf{METEOR} & \textbf{CIDEr}\\
\hline
\textbf{sc-vgg-16} & 30.1 & 24.7 & 97.0\\ 
\textbf{sc-resnet-34}  & 30.6 & 25.0 & \textbf{98.1}\\
\textbf{sc-resnet-200} & \textbf{31.4} & \textbf{25.1} & 97.7\\
\hline
\end{tabular}
\end{center}
\end{table}

\begin{table}[t]
\begin{center}
\caption{Top-6 nearest neighbors for randomly picked words. The results are based on the text-conditional matrix $W_c$ of 1-gram word-conditional model. We highlight similar words in semantics.}
\renewcommand\arraystretch{1.2}
\label{tbl:analysis}
\begin{tabular}{c|l} 
\hline
\textbf{dog} & \textbf{bear} \; three\; woman \; \textbf{cat} \; girl \; person \\
\hline
\textbf{banana} & it \; \textbf{carrots} \; \textbf{fruits} \; six \; onto \; includes \\
\hline
\textbf{red} & UNK \; \textbf{blue} \; three \; several \; man \; \textbf{yellow} \\
\hline
\textbf{sitting} & \textbf{standing} \; next \; are \; \textbf{sits} \; dog \; woman \\
\hline
\textbf{man} & \textbf{woman} \; \textbf{person} \; \textbf{his} \; three \; are \; dog \\
\hline
\end{tabular}
\end{center}
\end{table}

\subsubsection{Attention}

It is essential to verify whether our learned text-conditional attention is semantically meaningful. Each column in the text-conditional matrix $W_c$ is an attention mask for image features, and it corresponds to a word in our dictionary. It is expected that similar words should have similar masks (with some variations). To verify, we calculate the similarities among masks using Euclidean distance. We show five \textit{randomly} sampled words w.r.t. different parts of speech (noun, verb and adjective). Table~\ref{tbl:analysis} shows their top few nearest words. Most of the neighbors are related to the original word, and some of them are strongly related, such as ``cat'' for ``dog'', ``blue'' for ``red'', ``sits'' for ``sitting'', and ``woman'' for ``man''. This shows strong evidence that our model is learning meaningful text-conditional attention.

\subsection{Compare to State-of-The-Art Methods}
\label{sec:exp:compare}

We use LSTM with time-invariant image guidance (img-gLSTM)~\cite{JiGaFeICCV2015} and NeuralTalk2~\cite{NT2}, an implementation of~\cite{ViToBeCVPR2015}, as baselines. We also compare to a state-of-the-art non-attention-based model---LSTM with semantic embedding guidance (emb-gLSTM)~\cite{JiGaFeICCV2015}. Furthermore, we compare our method to a set of state-of-the-art attention-based methods including visual attention with soft- and hard-attention~\cite{XuBaKiICML2015}, and semantic attention with visual attributes (ATT-FCN)~\cite{YoJiWaCVPR2016}. For fair comparison among different attention models, we report our results with 16-layer VGGNet~\cite{SiZiICLR2015} since it is similar to the image encodings used in other methods. 

Table~\ref{tbl:result} shows the comparison results. Our methods, both 1-gram word-conditional and sentence-conditional, outperform our two baselines in all metrics by a large margin, ranging from 1\% to 5\%. The results are strong evidence that 1) our td-gLSTM is better suited for captioning comparing to time-invariant gLSTM; and 2) modeling textual context is essential for image captioning. Also, our methods yield much higher evaluation scores than emb-gLSTM~\cite{JiGaFeICCV2015} showing the effectiveness of using textual content in our model. 


We further compare our text-conditional methods with state-of-the-art attention-based methods. For 1-gram word-conditional method, the attention on the image feature guidance is merely determined by the previously generated word. Apparently, this results in semantic information loss. Even though, its performances are still on par with or better than state-of-the-art attention-based methods, such as Hard-Attention and ATT-FCN. We then upgrade the word-conditional model to the sentence-conditional model, which leads to improved performance in all metrics, and it outperforms state-of-the-art methods in most metrics. It worth noting that BLEU@1 score is related to single word accuracy, and highly affected by word vocabularies. This might result in our relatively low BLEU@1 score compared with hard-attention \cite{XuBaKiICML2015}.

\begin{table*}[t]
\begin{center}
\caption{Comparison to baselines and state-of-the-art methods. For some competing methods, we extract their performance from the corresponding papers. For a fair comparison, we use 16-layer VGGNet for image encoding. Top-two scores for each metric are highlighted. All values are reported as percentage (\%).} 
\renewcommand\arraystretch{1.1}
\label{tbl:result}  
\begin{tabular}{r|cccccc} 
\hline
\textbf{Methods} & \textbf{BLEU@1} & \textbf{BLEU@2} & \textbf{BLEU@3} & \textbf{BLEU@4} & \textbf{METEOR} & \textbf{CIDEr}\\
\hline
\textbf{img-gLSTM} \cite{JiGaFeICCV2015} &  64.7 & 45.9 & 31.1 & 21.4 & 20.4 & 67.7   \\
\textbf{emb-gLSTM} \cite{JiGaFeICCV2015} &  67.0 & 49.1 & 35.8 & 26.4 & 22.7 & 81.3   \\
\textbf{NeuralTalk2} \cite{NT2} &  70.5 & 53.2 & 39.2 & 28.9 & 24.3 & 92.3  \\
\hline
\textbf{Hard-Attention} \cite{XuBaKiICML2015} &  \textbf{71.8} & 50.4 & 35.7 & 25.0 & 23.0 & -   \\
\textbf{Soft-Attention} \cite{XuBaKiICML2015} &  70.7 & 49.2 & 34.4 & 24.3 & 23.9 & -   \\
\textbf{ATT-FCN} \cite{YoJiWaCVPR2016} &  70.9 & 53.7& \textbf{40.2} & \textbf{30.4} & 24.3 & -   \\
\hline
\textbf{Our 1-gram-vgg-16} & 71.5 & \textbf{54.2} & 40.0 & 29.6 & \textbf{24.5} & \textbf{95.5}\\
\textbf{Our sc-vgg-16} & \textbf{71.6} & \textbf{54.5} & \textbf{40.5} & \textbf{30.1} & \textbf{24.7} & \textbf{97.0}\\ 
\hline
\end{tabular}
\end{center}
\end{table*}

\begin{table}[t]
\begin{center}
\caption{Results of human evaluation. The higher the better for both the content quality and the grammar scores. The highest score for each column is highlighted.}
\vspace{10pt}
\label{tbl:he}  
\begin{tabular}{r|cc} 
\hline
\textbf{Methods} & \textbf{Content} & \textbf{Grammar}\\
\hline
\textbf{img-gLSTM} \cite{JiGaFeICCV2015} & 1.56 & 2.75 \\
\textbf{NeuralTalk2} \cite{NT2} & 1.94 & 2.77 \\
\textbf{Our sc-vgg-16} & \textbf{2.00} & \textbf{2.80} \\
\hline
\end{tabular}
\end{center}
\end{table}

\subsubsection{Human Evaluation}

We choose three methods for human evaluation, NeuralTalk2, img-gLSTM and our sentence-conditional attention model. A cohort of five well-trained human annotators is performed the experiments. Each of the annotators were shown 500 pairs of randomly selected images and three corresponding generated captions. The annotators rate the three captions from 0 to 3 regarding the content quality and grammar (the higher the better). For content quality, a score of 3 is given if the caption describes all the important content, e.g. objects and actions, in the image; a score of 0 is given if the caption is totally wrong or irrelevant. For grammar, a score of 3 denotes human-level natural expression and a score of 0 means the caption is unreadable. The results are shown in Table~\ref{tbl:he}. Our proposed sentence-conditional model lead the baseline img-gLSTM by a large margin of 28.2\% in the caption content quality, and 3.1\% compared to the baseline Neuraltalk2, showing the effectiveness of our attention mechanism in captioning. As for grammar, all the methods create human-like sentences with a few grammar mistakes, and adding sentence-conditional attention to LSTM yields a slightly higher grammar score, due to the explicitly textual information contained in the LSTM guidance input.

\begin{table*}[t]
\centering
\caption{Evaluation on MS-COCO leaderboard. We list state-of-the-art published results. We highlight our method and our baseline method, NeuralTalk2. Notice that methods highly-ranked in learderboard use better CNNs, inference methods, and more careful engineering than research publishes.}
\label{tbl:official}
\footnotesize
\begin{tabular}{|c|c|c|c|c|c|c|c|c|c|c|c|c|c|c|}
\hline
 & \multicolumn{2}{c|}{BLEU-1} & \multicolumn{2}{c|}{BLEU-2} & \multicolumn{2}{c|}{BLEU-3} & \multicolumn{2}{c|}{BLEU-4} & \multicolumn{2}{c|}{METEOR} & \multicolumn{2}{c|}{ROUGE-L} & \multicolumn{2}{c|}{CIDEr-D} \\
 \hline
Methods & c5 & c40 & c5 & c40 & c5 & c40 & c5 & c40 & c5 & c40 & c5 & c40 & c5 & c40 \\
\hline
\textbf{ATT\_VC} \cite{YoJiWaCVPR2016} & 0.731 & 0.900 & 0.565 & 0.815 & 0.424 & 0.709 & 0.316 & 0.599 & 0.250 & 0.335 & 0.535 & 0.682 & 0.943 & 0.958 \\
\hline
\textbf{OriolVinyals} \cite{vinyals2016show} & 0.713 & 0.895 & 0.542 & 0.802 & 0.407 & 0.694 & 0.309 & 0.587 & 0.254 & 0.346 & 0.530 & 0.682 & 0.943 & 0.946 \\
\hline
\textbf{jeffdonahue} \cite{DoAnGuCVPR2015} & 0.718 & 0.895 & 0.548 & 0.804 & 0.409 & 0.695 & 0.306 & 0.585 & 0.247 & 0.335 & 0.528 & 0.678 & 0.921 & 0.934 \\
\hline
\textbf{SC-Tanh (ours)} & \textbf{0.717} & \textbf{0.887} & \textbf{0.546} & \textbf{0.794} & \textbf{0.405} & \textbf{0.682} & \textbf{0.300} & \textbf{0.569} & \textbf{0.248 }&\textbf{0.330} & \textbf{0.515} &\textbf{ 0.667} & \textbf{0.923} & \textbf{0.929}\\
\hline
\textbf{Q.Wu} \cite{WuShLiCVPR2016} & 0.725 & 0.892 & 0.556 & 0.803 & 0.414 & 0.694 & 0.306 & 0.582 & 0.246 & 0.329 & 0.528& 0.672& 0.911 & 0.924\\
\hline
\textbf{Human} & 0.663	& 0.880 & 0.469 &	0.744 &	0.321 &	0.603 &	0.217 &	0.471 &	0.252 &	0.335 &	0.484 &	0.626 &	0.854 &	0.910\\
\hline
\textbf{NeuralTalk2} \cite{NT2} & \textbf{0.706}	& \textbf{0.877} & \textbf{0.530} &	\textbf{0.776} &	\textbf{0.388} &	\textbf{0.657} & \textbf{0.284} &	\textbf{0.541} &	\textbf{0.238} &	\textbf{0.317} &	\textbf{0.515} &\textbf{	0.654} &	\textbf{0.858} &	\textbf{0.865}\\
\hline
\end{tabular}
\end{table*}

\subsubsection{Qualitative Results} 

\begin{figure*}[t]
\centering
\includegraphics[width=\linewidth]{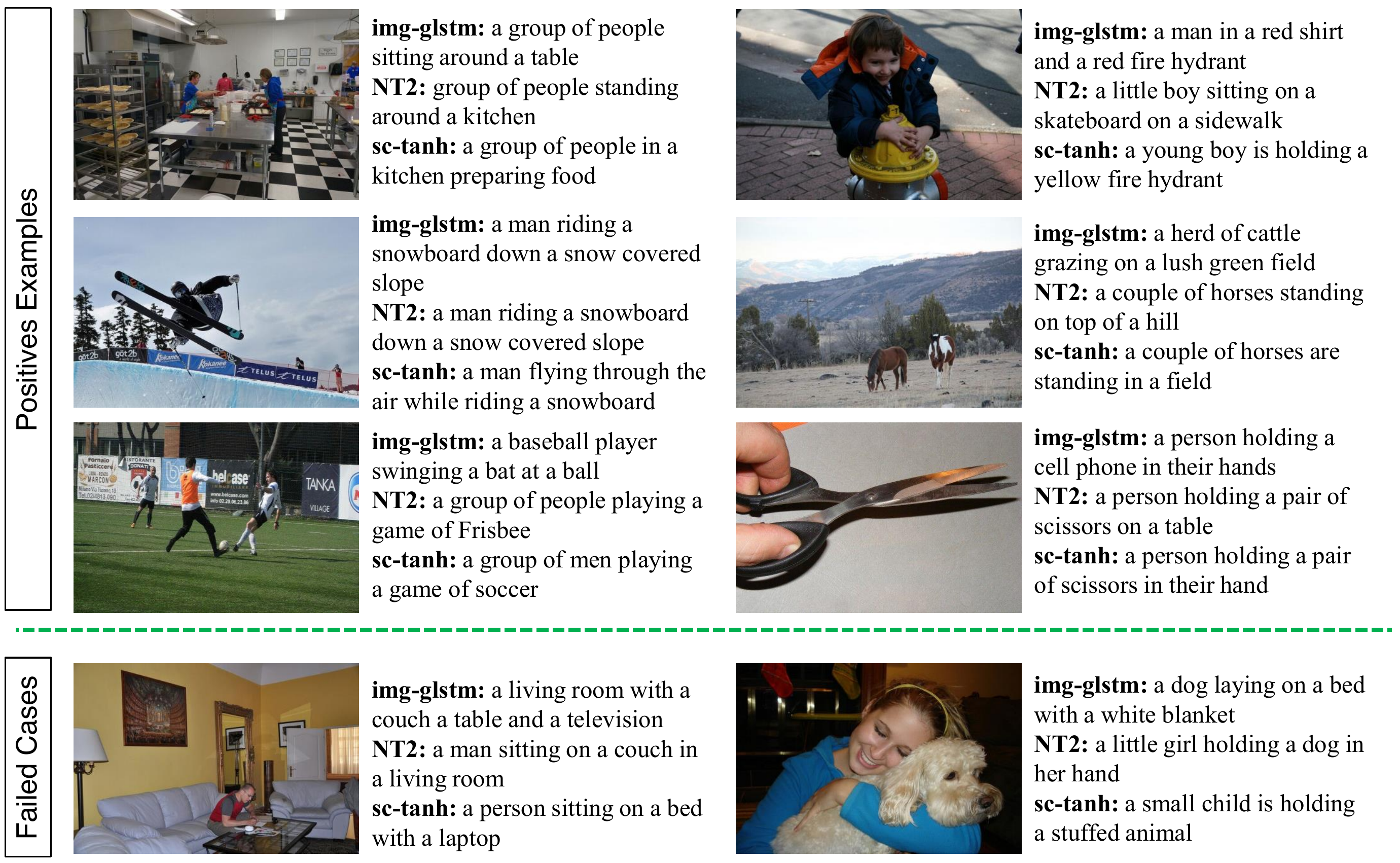}
\caption{Qualitative results. We show both positive examples and failed cases of our method. NT2 indicates NeuralTalk2 and \textbf{sc-tanh} is our sentence-conditional model. Better viewed in color.}
\label{fig:results}
\end{figure*}

Figure~\ref{fig:results} shows qualitative captioning results. The fix images in the first three rows are positive examples and the last two are failed cases. Our proposed model can better capture details in the target image, such as ``yellow fire hydrant'' in the second image, and ``soccer'' in the fifth image. Also, the text-conditional attention discovers rich context information in the image, such as the ``preparing food'' followed by ``kitchen'' in the first image, and the `in their hand`'' followed by ``holding'' in the sixth image. However, we also show the failed cases, where the objects are mistakenly inferred from the previous words. For the first image, when we feed in the word sequence ``a man (is) sitting'', our text-conditional attention is triggered by things can be sat by a man; a sofa is a reasonable candidate according to the training data. Similarly, for the second image, the model is trained on some images with stuffed animal held by a person, which in some sense biases the semantic attention model.

\subsubsection{Leaderboard Competition}

We test our model on the MS-COCO leaderboard competition and summarize the results in Table \ref{tbl:official}. Our method outperforms the baseline (NeuralTalk2 \cite{NT2}) across all the metrics and is on par with state-of-the-art methods. It worth noting that our baseline is an open source implementation of~\cite{ViToBeCVPR2015}, shown as OriolVinyals in Tab. \ref{tbl:official}, but the latter performs much better due to better CNNs, inference methods, and more careful engineering. Also, several methods unreasonably outperform human-annotated captions, which reveals the drawback of the existing evaluation metrics.


\section{Conclusion}

In this paper, we propose a semantic attention mechanism for image caption 
generation, called text-conditional semantic attention, which provides
explicitly text-conditioned image features for attention.
We also improve the existing gLSTM framework by introducing time-dependent
guidance, opening up a new way for further boosting image captioning
performance.
We show in our experiments that the proposed methods significantly improve
the baseline method and outperform state-of-the-art methods, which supports
our argument of explicit consideration of using text-conditional attention
modeling.

\noindent \textbf{Future Work.} There are several ways in which we can further improve our method. First, combining text-conditional attention with region-based or attribute-based attention, so that the model can learn to attend on regions in feature maps or attributes extracted from the image. Second, one common issue with supervised training is overfitting. As Vinyals et al.~\cite{ViToBeCVPR2015} pointed out, we cannot access
enough training samples, even for the relatively huge dataset such as MS-COCO. One possible solution is to combine weakly annotated images with current dataset, such as~\cite{GoWaHoECCV2014}. We keep those for our future work.

{\small
\bibliographystyle{ieee}
\bibliography{textcond}

\begin{thebibliography}{10}\itemsep=-1pt

\bibitem{AnVeRoCVPR2016}
L.~Anne~Hendricks, S.~Venugopalan, M.~Rohrbach, R.~Mooney, K.~Saenko, and
  T.~Darrell.
\newblock Deep compositional captioning: Describing novel object categories
  without paired training data.
\newblock In {\em IEEE Conference on Computer Vision and Pattern Recognition},
  2016.

\bibitem{BaChBeICLR2015}
D.~Bahdanau, K.~Cho, and Y.~Bengio.
\newblock Neural machine translation by jointly learning to align and
  translate.
\newblock In {\em International Conference on Learning Representations}, 2015.

\bibitem{ChVaGuEMNLP2014}
K.~Cho, B.~Van~Merri{\"e}nboer, C.~Gulcehre, D.~Bahdanau, F.~Bougares,
  H.~Schwenk, and Y.~Bengio.
\newblock Learning phrase representations using rnn encoder-decoder for
  statistical machine translation.
\newblock In {\em Empirical Methods in Natural Language Processing}, 2014.

\bibitem{DaSrCoWSDM2013}
P.~Das, R.~K. Srihari, and J.~J. Corso.
\newblock Translating related words to videos and back through latent topics.
\newblock In {\em WSDM}, 2013.

\bibitem{DeDuARN1995}
R.~Desimone and J.~Duncan.
\newblock Neural mechanisms of selective visual attention.
\newblock {\em Annual Review of Neuroscience}, 18(1):193--222, 1995.

\bibitem{DoAnGuCVPR2015}
J.~Donahue, L.~Anne~Hendricks, S.~Guadarrama, M.~Rohrbach, S.~Venugopalan,
  K.~Saenko, and T.~Darrell.
\newblock Long-term recurrent convolutional networks for visual recognition and
  description.
\newblock In {\em IEEE Conference on Computer Vision and Pattern Recognition},
  2015.

\bibitem{ElKeEMNLP2013}
D.~Elliott and F.~Keller.
\newblock Image description using visual dependency representations.
\newblock In {\em Empirical Methods in Natural Language Processing}, 2013.

\bibitem{FaGuIaCVPR2015}
H.~Fang, S.~Gupta, F.~Iandola, R.~K. Srivastava, L.~Deng, P.~Doll{\'a}r,
  J.~Gao, X.~He, M.~Mitchell, J.~C. Platt, et~al.
\newblock From captions to visual concepts and back.
\newblock In {\em IEEE Conference on Computer Vision and Pattern Recognition},
  2015.

\bibitem{FaHeSaECCV2010}
A.~Farhadi, M.~Hejrati, M.~Sadeghi, P.~Young, C.~Rashtchian, J.~Hockenmaier,
  and D.~Forsyth.
\newblock Every picture tells a story: Generating sentences from images.
\newblock In {\em European Conference on Computer Vision}, 2010.

\bibitem{GoWaHoECCV2014}
Y.~Gong, L.~Wang, M.~Hodosh, J.~Hockenmaier, and S.~Lazebnik.
\newblock Improving image-sentence embeddings using large weakly annotated
  photo collections.
\newblock In {\em European Conference on Computer Vision}, 2014.

\bibitem{graves2008super}
A.~Graves.
\newblock Supervised sequence labelling with recurrent neural networks.
\newblock 2012.

\bibitem{GuKrMaICCV2013}
S.~Guadarrama, N.~Krishnamoorthy, G.~Malkarnenkar, S.~Venugopalan, R.~Mooney,
  T.~Darrell, and K.~Saenko.
\newblock Youtube2text: Recognizing and describing arbitrary activities using
  semantic hierarchies and zero-shot recognition.
\newblock In {\em IEEE International Conference on Computer Vision}, 2013.

\bibitem{HeZhReCVPR2016}
K.~He, X.~Zhang, S.~Ren, and J.~Sun.
\newblock Deep residual learning for image recognition.
\newblock In {\em IEEE Conference on Computer Vision and Pattern Recognition},
  2016.

\bibitem{HoScNECO1997}
S.~Hochreiter and J.~Schmidhuber.
\newblock Long short-term memory.
\newblock {\em Neural Computation}, 9(8):1735--1780, 1997.

\bibitem{JiGaFeICCV2015}
X.~Jia, E.~Gavves, B.~Fernando, and T.~Tuytelaars.
\newblock Guiding the long-short term memory model for image caption
  generation.
\newblock In {\em IEEE International Conference on Computer Vision}, 2015.

\bibitem{JoKaFeCVPR2016}
J.~Johnson, A.~Karpathy, and L.~Fei-Fei.
\newblock Densecap: Fully convolutional localization networks for dense
  captioning.
\newblock In {\em IEEE Conference on Computer Vision and Pattern Recognition},
  2016.

\bibitem{NT2}
A.~Karpathy.
\newblock neuraltalk2.
\newblock \url{https://github.com/karpathy/neuraltalk2}, 2015.

\bibitem{KaFeCVPR2015}
A.~Karpathy and L.~Fei-Fei.
\newblock Deep visual-semantic alignments for generating image descriptions.
\newblock In {\em IEEE Conference on Computer Vision and Pattern Recognition},
  2015.

\bibitem{kingma2014adam}
D.~Kingma and J.~Ba.
\newblock Adam: A method for stochastic optimization.
\newblock {\em arXiv preprint arXiv:1412.6980}, 2014.

\bibitem{KuPrOrTPAMI2013}
G.~Kulkarni, V.~Premraj, V.~Ordonez, S.~Dhar, S.~Li, Y.~Choi, A.~C. Berg, and
  T.~L. Berg.
\newblock Babytalk: Understanding and generating simple image descriptions.
\newblock {\em IEEE Transactions on Pattern Analysis and Machine Intelligence},
  35(12):2891--2903, 2013.

\bibitem{KuOrBeACL2012}
P.~Kuznetsova, V.~Ordonez, A.~C. Berg, T.~L. Berg, and Y.~Choi.
\newblock Collective generation of natural image descriptions.
\newblock In {\em Association for Computational Linguistics}, 2012.

\bibitem{LiKuBeCNLL2011}
S.~Li, G.~Kulkarni, T.~L. Berg, A.~C. Berg, and Y.~Choi.
\newblock Composing simple image descriptions using web-scale n-grams.
\newblock In {\em Computational Natural Language Learning}, 2011.

\bibitem{LiMaBeECCV2014}
T.-Y. Lin, M.~Maire, S.~Belongie, J.~Hays, P.~Perona, D.~Ramanan,
  P.~Doll{\'a}r, and C.~L. Zitnick.
\newblock Microsoft coco: Common objects in context.
\newblock In {\em European Conference on Computer Vision}, 2014.

\bibitem{LiPaECCV2016}
X.~Lin and D.~Parikh.
\newblock Leveraging visual question answering for image-caption ranking.
\newblock In {\em European Conference on Computer Vision}, 2016.

\bibitem{MaXuYaICLR2015}
J.~Mao, W.~Xu, Y.~Yang, J.~Wang, Z.~Huang, and A.~Yuille.
\newblock Deep captioning with multimodal recurrent neural networks (m-rnn).
\newblock In {\em International Conference on Learning Representations}, 2015.

\bibitem{MiHaDoEACL2012}
M.~Mitchell, X.~Han, J.~Dodge, A.~Mensch, A.~Goyal, A.~Berg, K.~Yamaguchi,
  T.~Berg, K.~Stratos, and H.~Daum{\'e}~III.
\newblock Midge: Generating image descriptions from computer vision detections.
\newblock In {\em EACL}. Citeseer, 2012.

\bibitem{russakovsky2015imagenet}
O.~Russakovsky, J.~Deng, H.~Su, J.~Krause, S.~Satheesh, S.~Ma, Z.~Huang,
  A.~Karpathy, A.~Khosla, M.~Bernstein, et~al.
\newblock Imagenet large scale visual recognition challenge.
\newblock {\em International Journal of Computer Vision}, 115(3):211--252,
  2015.

\bibitem{SiZiICLR2015}
K.~Simonyan and A.~Zisserman.
\newblock Very deep convolutional networks for large-scale image recognition.
\newblock In {\em International Conference on Learning Representations}, 2015.

\bibitem{SuViLeNIPS2014}
I.~Sutskever, O.~Vinyals, and Q.~V. Le.
\newblock Sequence to sequence learning with neural networks.
\newblock In {\em Advances in Neural Information Processing Systems}, 2014.

\bibitem{vedantam2015cider}
R.~Vedantam, C.~Lawrence~Zitnick, and D.~Parikh.
\newblock Cider: Consensus-based image description evaluation.
\newblock In {\em Proceedings of the IEEE Conference on Computer Vision and
  Pattern Recognition}, pages 4566--4575, 2015.

\bibitem{ViToBeCVPR2015}
O.~Vinyals, A.~Toshev, S.~Bengio, and D.~Erhan.
\newblock Show and tell: A neural image caption generator.
\newblock In {\em IEEE Conference on Computer Vision and Pattern Recognition},
  2015.

\bibitem{vinyals2016show}
O.~Vinyals, A.~Toshev, S.~Bengio, and D.~Erhan.
\newblock Show and tell: Lessons learned from the 2015 mscoco image captioning
  challenge.
\newblock {\em IEEE Transactions on Pattern Analysis and Machine Intelligence},
  2016.

\bibitem{WuShLiCVPR2016}
Q.~Wu, C.~Shen, L.~Liu, A.~Dick, and A.~v.~d. Hengel.
\newblock What value do explicit high level concepts have in vision to language
  problems?
\newblock In {\em IEEE Conference on Computer Vision and Pattern Recognition},
  2016.

\bibitem{XuBaKiICML2015}
K.~Xu, J.~Ba, R.~Kiros, K.~Cho, A.~Courville, R.~Salakhutdinov, R.~S. Zemel,
  and Y.~Bengio.
\newblock Show, attend and tell: Neural image caption generation with visual
  attention.
\newblock In {\em International Conference on Machine Learning}, 2015.

\bibitem{YaTeDaEMNLP2011}
Y.~Yang, C.~L. Teo, H.~Daum{\'e}~III, and Y.~Aloimonos.
\newblock Corpus-guided sentence generation of natural images.
\newblock In {\em Proceedings of the Conference on Empirical Methods in Natural
  Language Processing}, pages 444--454. Association for Computational
  Linguistics, 2011.

\bibitem{YoJiWaCVPR2016}
Q.~You, H.~Jin, Z.~Wang, C.~Fang, and J.~Luo.
\newblock Image captioning with semantic attention.
\newblock In {\em IEEE Conference on Computer Vision and Pattern Recognition},
  2016.

\end{thebibliography}
}

\end{document}